\documentclass[letterpaper, 10 pt, conference]{ieeeconf}  

\IEEEoverridecommandlockouts                              

\usepackage{amsmath,amssymb,amsfonts}
\usepackage{algorithmic}
\usepackage{graphicx}
\usepackage{textcomp}
\usepackage{comment}
\usepackage{xcolor}
\usepackage{multirow}
\usepackage{booktabs}
\usepackage{amsfonts,amssymb}
\usepackage{cite}
\usepackage{url}
\usepackage{balance}

\usepackage{caption}
\usepackage{subcaption}
\usepackage{physics}
\title{\LARGE \bf
Correlated Time Series Self-Supervised Representation Learning via Spatiotemporal Bootstrapping
}

\author{Luxuan Wang$^{1}$, Lei Bai$^{2}$, Ziyue Li$^{3}$, Rui Zhao$^{4}$, Fugee Tsung$^{5}$
\thanks{$^{1}$Luxuan Wang is with Interdisciplinary Programs Office, The Hong Kong University of Science and Technology, Hong Kong SAR (Email: lwangda@connect.ust.hk)
        }%
\thanks{$^{2}$Lei Bai is with The Shanghai AI Laboratory, Shanghai, China (Email: baisanshi@gmail.com)}%
\thanks{$^{3}$Ziyue Li is with Information System, University of Cologne, 50923 Cologne, NRW, Germany (Email: zlibn@wiso.uni-koeln.de)}%
\thanks{$^{4}$Rui Zhao is with SenseTime Reasearch and Qing Yuan Research Institute of Shanghai Jiao Tong University, Shanghai, China (Email: zhaorui@sensetime.com)}%
\thanks{$^{5}$Fugee Tsung is with The Hong Kong University of Science and Technology and The Hong Kong University of Science and Technology (Guangzhou), Guangzhou, China (Email:season@ust.hk)}%
}

\begin{document}
\maketitle
\thispagestyle{empty}
\pagestyle{empty}

\begin{abstract}
Correlated time series analysis plays an important role in many real-world industries. Learning an efficient representation of this large-scale data for further downstream tasks is necessary but challenging. In this paper, we propose a time-step-level representation learning framework for individual instances via bootstrapped spatiotemporal representation prediction. We evaluated the effectiveness and flexibility of our representation learning framework on correlated time series forecasting and cold-start transferring the forecasting model to new instances with limited data. A linear regression model trained on top of the learned representations demonstrates our model performs best in most cases. Especially compared to representation learning models, we reduce the RMSE, MAE, and MAPE by 37$\%$, 49$\%$, and 48$\%$ on the PeMS-BAY dataset, respectively. Furthermore, in real-world metro passenger flow data, our framework demonstrates the ability to transfer to infer future information of new cold-start instances, with gains of 15$\%$, 19$\%$, and 18$\%$. The source code will be released under the GitHub \url{https://github.com/bonaldli/Spatiotemporal-TS-Representation-Learning}.
\end{abstract}


\section{Introduction}
Correlated time series (CTS) exists extensively in many real-world industries, including traffic management \cite{li2020tensor,li2020long,lin2023dynamic}, climate science \cite{cirstea2018correlated,hu2018risk,guo2014towards}, and so on. It refers to a set of time series that are not independent but have correlations between instances. For example, in Intelligent Transportation System (ITS) \cite{figueiredo2001towards,li2022individualized,lan2023mm,ziyue2021tensor,jiang2023unified}, sensors are installed at various locations to detect the traffic speed and flow on the road. The traffic time series are often correlated because traffic on one road is often correlated with traffic on neighboring roadways \cite{pedersen2020anytime,li2020tensor}. For example, as shown in Fig. \ref{example}(a), when two instances are quite geographically close, their attributes (e.g., traffic flow) are also quite similar. With the continued development of technologies in the Internet of Things (IoT), large amounts of correlated time series are produced every day. Learning an efficient representation of this large-scale data for further downstream tasks is necessary but challenging. In recent years, inspired by the great success of self-supervised representation learning in computer vision and natural language process, many studies \cite{ijcai2021-324,franceschi2019unsupervised,Yue2021TS2VecTU} have proposed various methods to learn latent representations of time series, and contrastive methods are currently the state-of-art approaches among discriminative self-supervised learning methods. However, the current approaches still have several significant shortcomings:

\begin{figure}[t]
\centerline{\includegraphics[width=0.48\textwidth]{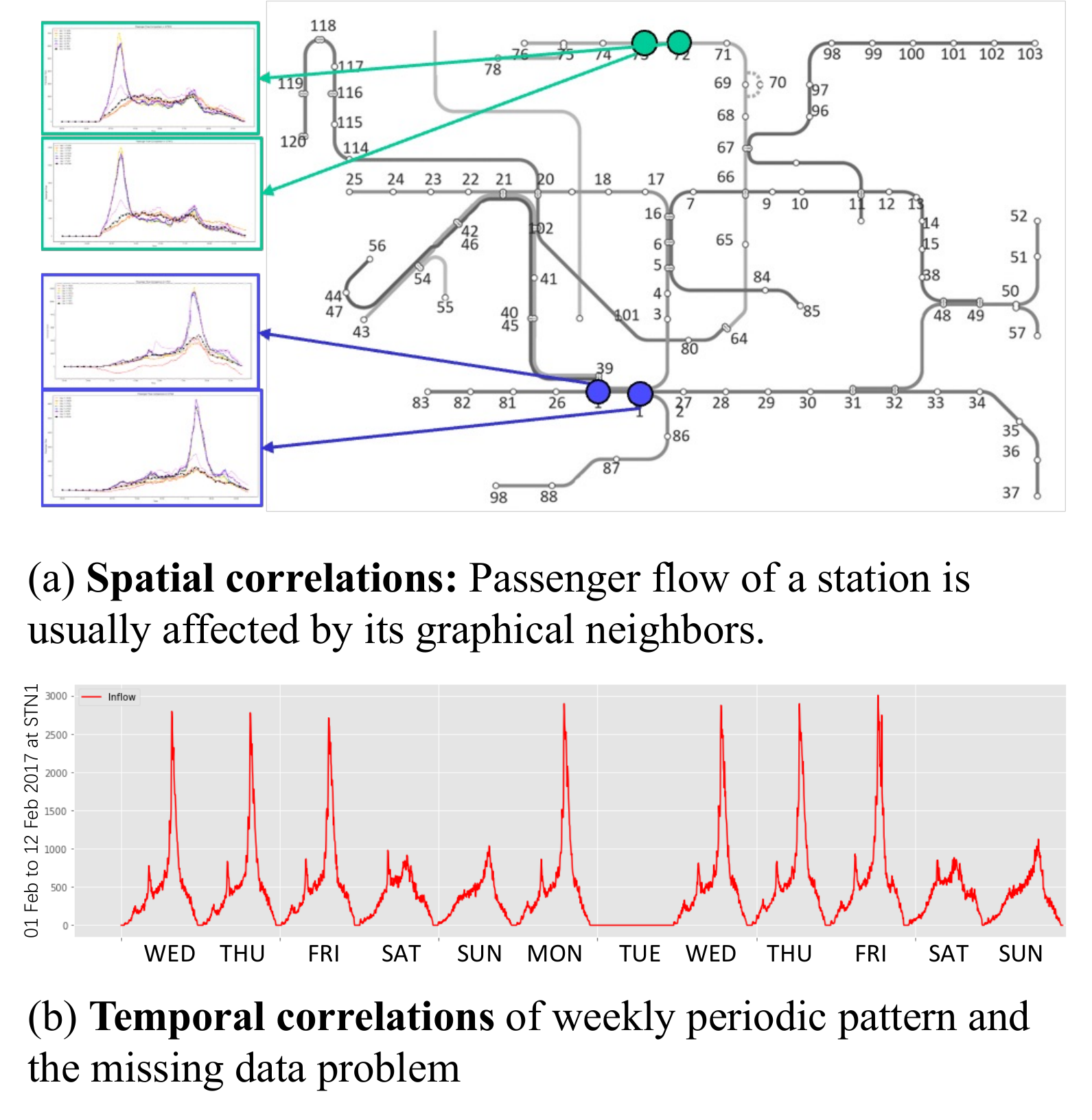}}
\caption{An example of the spatial and temporal correlations of correlated time series: passenger inflow of metro stations \cite{li2020tensor,li2020long}. (a) \textbf{Spatial  Correlation:} Passenger flows of neighboring stations are usually similar. (b) \textbf{Temporal  Correlation:} Passenger flow of a station usually has historical period patterns. Besides, missing data is also observed.}
\label{example}
\end{figure}

Firstly, most recent studies \cite{ijcai2021-324,franceschi2019unsupervised} only learn instance-level representations, which is unsuitable for point-wise tasks, e.g., forecasting and anomaly detection. These studies demonstrated success in many time series downstream tasks, especially for classification and clustering. Unfortunately, the representation of the whole time series could not be adapted to infer specific time steps or sub-series.

Secondly, current studies neglect the correlation between different instances and learn an integrated representation for correlated (multivariate) time series, which is impossible to transfer to some real-world downstream tasks, e.g., sub-instance forecasting (explained in detail in our downstream tasks). For example, TS2Vec \cite{Yue2021TS2VecTU} performs contrastive learning over augmented context views in a hierarchical manner and leanings a robust contextual representation for each timestamp. However, when dealing with multivariate cases, TS2Vec only learns an aggregated representation of all instances and does not consider the correlations between instances.

Thirdly, all the contrast-based methods \cite{tonekaboni2021unsupervised,franceschi2019unsupervised,Yue2021TS2VecTU,mao2022jointly} construct positive pairs or negative samples based on prior knowledge of the time series dataset or make strong assumptions of the data distribution of time series. These approaches may lose the variety of positive pairs due to the heavy overlap between positive pairs. Furthermore, since periodicity is a common phenomenon in time series data, as shown in Fig. \ref{example}(b), the above contrasting methods often suffer the problem of false negative samples. 

To tackle these challenges, we proposed a correlated time series representation learning framework via spatiotemporal bootstrapping. The major contributions of this paper are summarized as follows:
\begin{itemize}
\item We propose a representation learning framework to learn point-level representation for arbitrary instances at any time step. With flexibility, the learned representations can be fine-tuned to 1) correlated time series forecasting tasks as well as 2) transfer tasks for the new instances without retraining the forecasting model. To capture both spatial and temporal correlations, we leverage historical data and neighborhood information from the predefined adjacent matrix in our self-supervised learning framework.
\item  To avoid frequent false negatives in contrastive learning paradigm for time series data, we incorporate a self-supervised learning framework without negative samples into our model. We define spatial and temporal targets and use a masked view to predict the representation of the targets to learn the correlations. 
\item  We demonstrate the effectiveness and flexibility of our representation learning framework based on results and analysis of different downstream tasks on real-world data. Our evaluation of the forecasting task shows comparable performance end-to-end solutions, but our method is more flexible to transfer to new instances entering the dataset without retraining the model.

\end{itemize}

The rest of the paper is organized as follows: Section \ref{sec: review} first reviews the related work of time series representation learning from two perspectives of pretext task learning and contrastive learning. Section \ref{sec: method} then give the detailed model. Section \ref{sec: experiment} conducted experiments based on two tasks of forecasting and cold-start transferring. Section \ref{sec: analysis} gives prudent analysis based on ablation studies and model robustness analysis. Section \ref{sec: conclusion} gives the conclusion and future work. 

 \begin{figure*}[t]
\centerline{\includegraphics[width=0.99\textwidth]{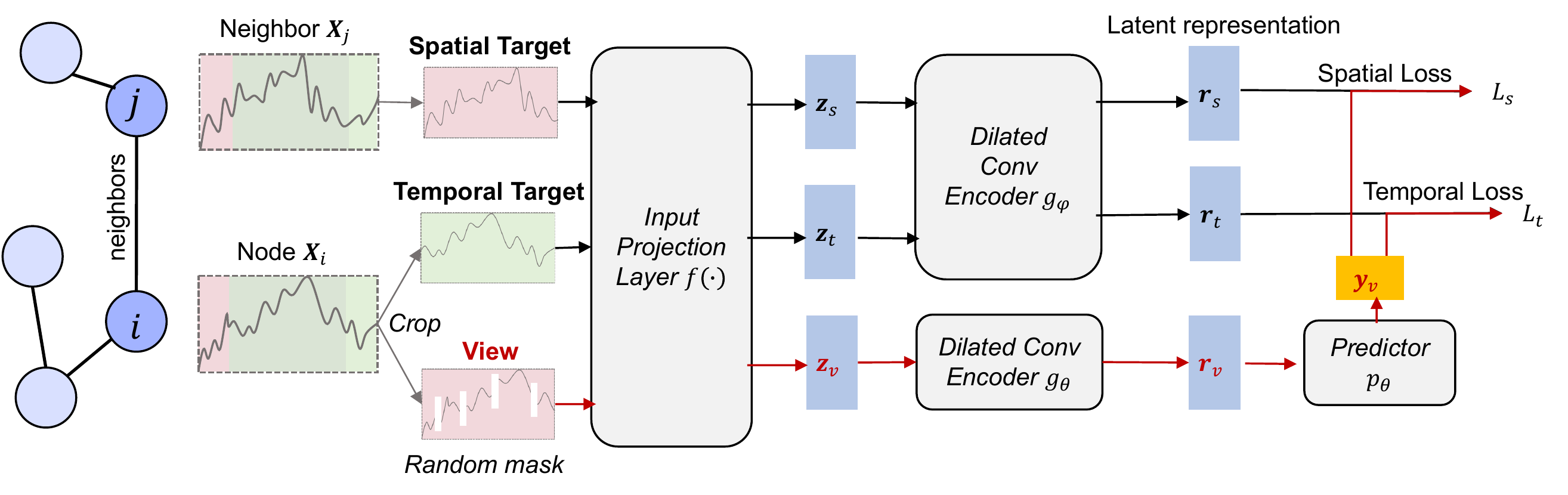}}
\caption{The proposed representation learning framework for correlated time series: The backbones consist of a linear layer projecting the series into a high-dimensional space, two dilated convolution encoders with different parameters, and a predictor to predict the spatiotemporal representations from the latent vector embedded from the masked view.}
\label{framework}
\end{figure*}

\section{Related Work}
\label{sec: review}
\subsection{Pretext Task Time Series Representation Learning}
A group of time series representation learning designs various pretext tasks to extract useful information to embeddings assist the model to perform well on downstream tasks. A typical pretext task is the reconstruction of masked series. Some earlier \cite{8621542} works apply autoencoders to reconstruct the input series to extract the latent representation by the encoder. Recent work such as  TST\cite{zerveas2021transformer} uses a Transformer backbone to reconstruct masked series by setting masked points to be zero to exact semantic-level representations. However, the masking technique changes the original data distribution, and since the time series often contains low-dense information, point-level masking only covers little information that needs to be recovered.  Another work \cite{banville2021uncovering} designs three pretext tasks, including two non-contrastive tasks: relative positioning, temporal shuffling and a contrastive  predictive coding task on EEG data to learn the representation. The first task is to measure the relative position between a specific window and other windows, whereas the second task refers to distinguishing the temporal order of shuffled windows. Rocket \cite{dempster2020rocket} incorporates a linear classifier on the top of embedding to learn high-level semantic representations.  

\subsection{Contrast-based Time Series Representation Learning}
In recent years, most time series representation learning incorporates contrastive learning framework \cite{hyvarinen2016unsupervised,chen2020simple}. Positive and negative pairs are formed using contrastive-based self-supervised learning techniques, where the contrastive loss attempt to maximize the  similarity between positive pairs while minimizing the similarity between negative pairs. Prior research suggests many methods for selecting positive and negative pairs to raise the standard of the learned representations. One of the earliest works, T-Loss \cite{franceschi2019unsupervised}, following Word2Vec \cite{mikolov2013efficient}, attempts to learn scalable representations by randomly selecting time segments via triplet loss. To investigate sub-series consistency, it treats a sub-series from the input time segment as a positive sample and randomly selects negative samples from other occurrences. Another earlier work, i.e., Contrastive Predictive Coding (CPC) \cite{oord2018representation} focuses on temporal series (including voice signal analysis) and learns the temporal correlations by splitting the entire series into consecutive windows and using an autoregressive model to predict the future window by the latent representation vector of the past window via Noise-Contrastive Estimation \cite{gutmann2010noise} loss. Some works were proposed to improve CPC. For instance, ContrNP \cite{kallidromitis2021contrastive} replaces the autoregressive model used by CPC with a neural network. Another following work is Temporal and Contextual Contrasting (TS-TCC) \cite{eldele2021time}, which learns embeddings for each input signal via a prediction task that uses the past part of each view to predict the future part of the other view through tough cross-view with strong and weak augmentations. Temporal Neighborhood Coding (TNC) \cite{tonekaboni2021unsupervised} instead introduces the concept of local stationary properties of signal data and assumes neighboring windows can be defined as positive samples while signals outside of this neighborhood are considered negative samples. It also introduces a sample weight adjustment technique to deal with non-stationary multivariate time series.  TS2Vec \cite{Yue2021TS2VecTU}, performing data augmentation of cropping and random masking on the projected high dimension space and considers augmented views from the same time step to be positive, whereas views from different time steps are considered to be negatives. It also proposes the instance-wise contrast between samples in the same batch. Their main difference in existing works is the policy and assumption of constructing contrastive pairs according. However, they often neglect the correlations between instances and suffer the problem of false negatives. 
 
\section{Methodology}
\label{sec: method}
We first give the preliminaries and definitions about correlated time series representation learning, which are necessary to formalize the problem. Throughout this exposition, scalars are denoted in italics, e.g., $n$; vectors and matrices by boldface letters, e.g., $\mathbf{X}$; Sets by boldface script capital $ \boldsymbol{\mathcal{X}}$.

\subsection{Preliminaries}
A time series $\mathbf{X}_i = \left\{x_{i,1}, x_{i,2}, \cdots, x_{i,T} \right\} $ is a temporal sequence of measurements, where the measurement $x_{i, t}$ is recorded at timestamp $t$ from the $i$-th instance, and $T$ is the total time scope. A \textbf{correlated} time series set is denoted as $\mathcal{X} = \{\mathbf{X}_1, \mathbf{X}_2, \cdots, \mathbf{X}_N\}$, where the time-series from different instances $X_i$ are spatially correlated with each other. To model spatial correlations among different time series, we introduce a graph $\mathcal{G} = (\mathcal{X}, \mathbf{A})$, where each instance in $\mathcal{X}$ corresponds to a single time series so that $|\mathcal{X}| = N$, and adjacency matrix $\mathbf{A} \in \mathbb{R}^{N \times N}$ contains weights to describe the strengths of the spatial correlations between each single time series. For example, in the traffic flow example, each $X_i$ represents the time-ordered flow recorded by each sensor, respectively, with a pre-defined distance-based adjacent matrix to describe the spatial correlations.

\subsection{Problem Definition}
Given a set of correlated time series $\mathcal{X} = \{\mathbf{X}_1, \mathbf{X}_2, \cdots, \mathbf{X}_N\}$ of $N$ instances, the goal is to learn a nonlinear embedding function $f(\cdot)$ that maps each $\mathbf{X}_i$ to its point-level representation $r_i$ that best describes itself. The representation $\mathbf{R}_i = \left\{\mathbf{r}_{i,1}, \mathbf{r}_{i,2}, \cdots, \mathbf{r}_{i,T}\right\}$ contains representation vectors $\mathbf{r}_{i,t} \in \mathbb{R}^K$ for each timestamp $t$ for each instance, where $K$ is the dimension of representation vectors.

\subsection{Construction of View and Targets}
For a sliding sub-window [$t, t+L_0$] in one instance $ \mathbf{X}_i \in {\mathbb{R}^{{1}\times{T}}}$ (single time series), we randomly sample two overlapping sub-series with the same window size $L_1$, starting at time $t$, and $t+l$ respectively. $L_0$, $L_1$, and $l$ are predefined length with $l<L_1<L_0$ and $L_0 = l + L_1$. Then we random mask some time steps of the cropped sub-series $w_{i,\;t}$ starting at $T=t$ and name it as view. Specifically, we mask the view $w_{i,\;t}$ with some continuous masks. For another cropped sub-series $w_{i,\;t+l}$, we name it a temporal target to emphasize the temporal consistency of the view. Based on the adjacent matrix, we sample the neighbor of $x_i$, which is notated as $x_j$. Similarly, we crop the sub-series $w_{j,\;t}$ at time $t$ with window size $L_1$ as the spatial target.

\subsection{Model architecture}
The encoder consists of three components: an input projection layer, a dilated casual CNN module, and an MLP predictor. For each input $w$, the input projection layer is a fully connected layer that maps the observation $x_{i,t}$ at timestamp $t$ to a high-dimensional latent vector $\mathbf{z}$. A dilated casual CNN module \cite{yu2015multi} with ten residual blocks is then applied to extract the contextual representation at each timestamp. Note the view and targets do not share the same parameters as the dilated casual CNN module. Each block of dilated casual CNN contains two 1-D convolutional layers with a dilation parameter ($2^l$ for the $l$-th block). The dilated casual convolutions enable a large receptive field for different domains \cite{bai2018empirical}.

Following the idea of BYOL\cite{grill2020bootstrap}, which learns its
representation by predicting the representation embedded from different augmented views of itself without using negative pairs, we learn representations by predicting different targets of the view instead of contrast. After view and targets are mapped to the high-dimensional latent vectors by the fully-connected layer $f$, we use two dilated casual neural networks to learn: the online $g_{\theta}$ and target networks $g_{\varphi}$. The two convolution networks have the same architectures but use different sets of weights $\theta$ and $\phi$, respectively. Then, the MLP predictor $p_{\theta}$ uses the learned representation of the view $r_v$ to predict the representations of targets in a joint manner. The parameter of $g_{\theta}$ and $p_{\theta}$ are updated by gradient while the parameter of $g_{\varphi}$ is updated by an exponential moving average algorithm to avoid model collapse\cite{grill2020bootstrap}. 

To calculate the spatiotemporal self-supervised loss, given a view $w_{i,\;t}$ and its temporal target $w_{i,\;t+l}$, spatial target $w_{j,\;t}$, our model produces their high-dimensional mapping vector $\mathbf{z}^{view} = f(w_{i,\;t})$, $\mathbf{z}^{t} = f(w_{i,\;t+l})$, and $\mathbf{z}^{s} = f(w_{j,\;t})$. Then, the online dilated casual convolution network outputs the latent representation of the view $\mathbf{r}_{\theta}^{view} = {g_{\theta}(\mathbf{z}^{view})}$ and a prediction $\mathbf{y}_{\theta}^{view} = p_{\theta}(\mathbf{r}_{\theta}^{view})$. The target network outputs the representations of the targets $\mathbf{r}_{\varphi}^t=g_{\varphi}(\mathbf{z}_{\varphi}^t)$ and $\mathbf{r}_{\varphi}^s =g_{\varphi}(\mathbf{z}_{\varphi}^s)$. Then we output the $l_2$-normalized prediction $\overline{\mathbf{y}}_{\theta}^{view} = \mathbf{y}_{\theta}^{view}/\norm{\mathbf{y}_{\theta}^{view}}_2$ and target representations $\overline{\mathbf{r}}_{\varphi}^t = \mathbf{r}_{\varphi}^t/\norm{\mathbf{r}_{\varphi}^t}_2$, $\overline{\mathbf{r}}_{\varphi}^s = \mathbf{r}_{\varphi}^s/\norm{\mathbf{r}_{\varphi}^s}_2$.
Finally, we define the temporal loss in Eq. (\ref{temporal loss}) and spatial loss in Eq. (\ref{spatial loss}) as following:  

\begin{equation}
L_{\theta,\varphi}^t = \norm{\overline{\mathbf{y}}_{\theta}^{view}-\overline{\mathbf{r}}_{\varphi}^t}^2_2 = 2-\frac{2\cdot \langle \mathbf{y}_{\theta}^{view}, \; \mathbf{r}_{\varphi}^t\rangle}{\norm{\mathbf{y}_{\theta}^{view}}_2 \cdot \norm{\mathbf{r}_{\varphi}^t}_2}
\label{temporal loss}
\end{equation}

\begin{equation}
L_{\theta,\varphi}^s = \norm{\overline{\mathbf{y}}_{\theta}^{view}-\overline{\mathbf{r}}_{\varphi}^s}^2_2 = 2-\frac{2\cdot \langle \mathbf{y}_{\theta}^{view}, \; \mathbf{r}_{\varphi}^s\rangle}{\norm{\mathbf{y}_{\theta}^{view}}_2 \cdot \norm{\mathbf{r}_{\varphi}^s}_2}
\label{spatial loss}
\end{equation}

The final loss function is defined as
\begin{equation}
L_{\theta, \varphi}^{s,t} = \alpha \; L_{\theta,\varphi}^t  + (1-\alpha)\; L_{\theta,\varphi}^s 
\label{loss}
\end{equation}

\noindent where the $\alpha$ is the tuning parameter to adjust the relative effect of spatial contrast and temporal contrast. At each training step, we perform a stochastic optimization step to minimize $L^{s,t}$ with respect to $\theta$ only, but not $\varphi$. The parameter updating of the proposed method is given as:

\begin{equation}
    \theta \gets optimizer(\theta, \nabla_{\theta} L_{\theta, \varphi}^{s,t}, \eta)
\end{equation}
\begin{equation}
    \varphi \gets \tau \varphi + (1-\tau) \theta
\end{equation}
where $optimizer(\cdot)$ is the Adam optimizer \cite{kingma2014adam} and $\eta$ is a learning rate. At the end of the training, we only keep the encoder $g_{\theta}$ for downstream tasks.
The model framework is summarized in Figure \ref{framework}.

\section{Experiments}
\label{sec: experiment}
In this section, we evaluate our learned representations on the following two downstream tasks: correlated time series forecasting and cold-start transferring forecasting model to new instances with limited data.

\subsection{Correlated Time Series Forecasting}
\begin{table*}[htbp]
	\centering
	\fontsize{8}{11}\selectfont    
	\caption{MAE, MAPE and RMSE for proposed method and the baselines. The best results are bolded.}
	\begin{tabular}{cccccccc}
		\toprule
		\toprule
		\multicolumn{2}{c}{\multirow{2}{*}{Model}}&
		\multicolumn{3}{c}{METR-LA Dataset (15/30/60 min)}& \multicolumn{3}{c}{PEMS-BAY Dataset (15/30/60 min)}\cr
		\cmidrule(lr){3-8}
		& & RMSE & MAE & MAPE(\%) & RMSE & MAE & MAPE(\%) \cr
		\cmidrule(lr){1-8}
		\multirow{5}{*}{End-to-end Model} &
		HA & 10.00 & 4.79 & 13.00 & 5.59 & 2.88 & 6.8   \cr
		\multirow{5}{*}{} & ARIMA & 8.21/10.45/13.23 & 3.99/5.15/6.90 & 9.60/12.70/17.40 & 3.30/4.76/6.50 & 1.62/2.33/3.38 & 3.50/5.40/8.30 \cr
		\multirow{5}{*}{} & VAR & 7.8/9.13/10.11 & 4.42/5.41/6.52 & 13.00/12.70/15.80 & 3.16/4.25/5.44 & 1.74/2.32/2.93 & 3.60/5.00/6.50 \cr
		\multirow{5}{*}{} & SVR & 8.45/10.87/13.76 & 3.99/5.05/6.72 & 9.30/12.10/16.70 & 3.59/5.18/7.08 & 1.85/2.48/3.28 & 3.80/5.50/8.00\cr
		\multirow{5}{*}{} & FC-LSTM & \textbf{6.30/7.23/8.69} & 3.44/\textbf{3.77/4.37} & 9.60/10.09/14.00 & 4.19/4.55/4.96 & 2.05/2.20/2.37 & 4.80/5.20/5.70 \cr
       \cmidrule(lr){1-8}
		\multirow{2}{*}{Fine-tuning Model} & TS2Vec & 8.92/9.65/10.44 & 6.00/6.34/6.93 & 13.15/14.04/15.95 & 5.25/5.40/5.74 & 3.27/3.33/3.49 & 6.72/6.90/7.33\cr
		\multirow{2}{*}{} & Proposed & 6.63/7.51/9.02 & \textbf{3.31}/3.80/4.77 & \textbf{7.80/9.44/12.23} & \textbf{2.59/3.54/4.78} & \textbf{1.27/1.64/2.21} & \textbf{2.63/3.65/5.28} \cr
		
		\bottomrule
		\bottomrule
	\end{tabular}\vspace{0cm}
	\label{tab:Training_sizes}
\end{table*}

Given a correlated time series set $\mathcal{X}_{1,2,\cdots,t} = \left\{ {\mathbf{X}_1, \mathbf{X}_2, \cdots, \mathbf{X}_N} \right\}$, we aim at predicting the future measurements of a specific, target time series in $\mathcal{X}$, i.e., $\mathbf{X}_{t+1, t+2,…, t+p}$, where $p$ is the prediction horizon. In our cases, the training phase consists of two steps: 1) learning time series representations $\mathbf{r}_{i,t}$, and 2) training a linear regression model on the representation of at time step $t$ and the future true value $\mathbf{X}_{t+1, t+2,…, t+p}$ with horizon $p$. The inference phase also consists of two steps: 1) inference of representations for associated timestamps and 2) forecasting using a trained linear regression model. It is worth mentioning that the representation model only needs to be trained once to account for various horizon settings.

We conduct experiments on two real-world traffic flow datasets:

\begin{itemize}
\item \textbf{METR-LA.} This traffic dataset contains traffic flow records collected from loop detectors in the highway of Los Angeles County \cite{jagadish2014big}. There are four months of data ranging from Mar 1st, 2012 to Jun 30th, 2012 collected at 207 sensors for the experiment. 
\item \textbf{PEMS-BAY.} This traffic dataset is collected by California Transportation Agencies (CalTrans) Performance Measurement System (PeMS). There are three months of data ranging from Jan 1st, 2017, to May 31st, 2017, collected at 325 sensors for the experiment. 
\end{itemize}

For the forecasting task, we use Root Mean Squared Errors (RMSE), Mean Absolute Errors (MAE), and Mean Absolute Percentage Errors (MAPE) as the evaluation metrics. We compare our framework with the following methods in two categories: 
\begin{itemize}
\item \textbf{End-to-end Model.}
1) \textbf{Historical Average} (HA); 2) \textbf{Auto-Regressive Integrated Moving Average} (ARIMA); 3) \textbf{VAR}: Vector Auto-Regression \cite{hamilton1994time}; 4) \textbf{SVR}: Support Vector Regression which uses a linear support vector machine for the regression task; 5) \textbf{Recurrent Neural Network} with fully connected LSTM hidden units (FC-LSTM) \cite{sutskever2014sequence};
\item \textbf{Pre-training and Fine-tuning.}
6). \textbf{TS2Vec} \cite{Yue2021TS2VecTU}. As one of the most state-of-the-art methods, this approach follows the paradigm that learns a robust representation first, then uses a linear regression model to fine-tune. 
\end{itemize}

The evaluation results on RMSE, MAE, and MAPE for METR-LA and PEMS-BAY are shown in Table I. In general, our method achieves a 20$\%$ average decrease of RMSE, 39$\%$ of MAE, 32$\%$ of MAPE on METR-LA and a 37$\%$ average decrease of RMSE, 49$\%$ of MAE, 48$\%$ of MAPE on PEMS-BAY compared to TS2Vec, indicating our point-level correlated time series representation learning framework is more efficient than TS2Vec to capture the spatiotemporal correlations. 

\begin{figure}[htbp]
\centerline{\includegraphics[width=0.49\textwidth]{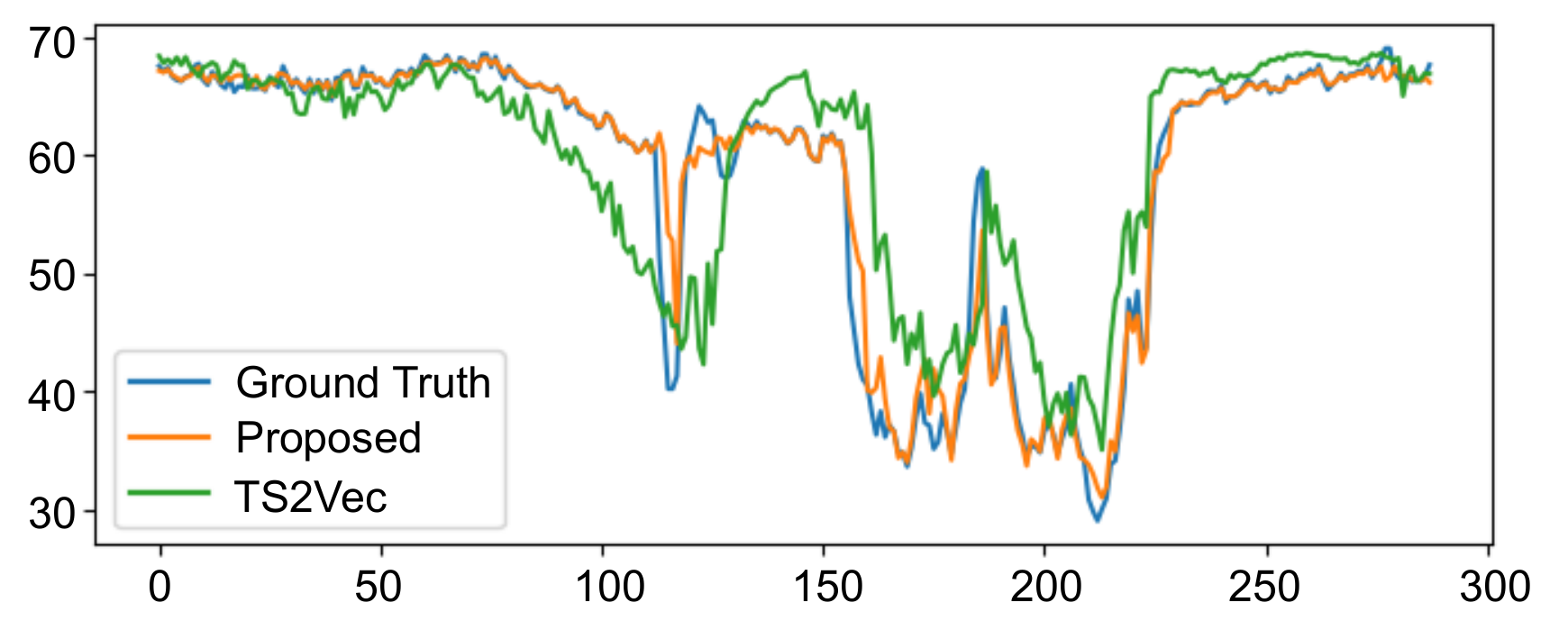}}
\caption{Visualization of a typical forecasting slice of our model and TS2Vec. Our model forecasts well on abrupt changes.}
\label{forecasting}
\end{figure}

Furthermore, our performance on the PEMS-BAY dataset is better than spatiotemporal graph-based neural networks for short-term forecasting on the PEMS-BAY dataset. Figure \ref{forecasting} presents a typical forecasting slice of our method and TS2Vec. We can see that, compared to TS2Vec, our model is more likely to predict rapid changes in traffic speed accurately. This is because our method takes advantage of spatial dependence and can exploit speed variations in nearby sensors for more precise predicting.

\subsection{Cold-Start Transferring Model to New Instances with Limited Data}

Considering a set of correlated time series X with original graph $\mathcal{G} = (\mathcal{X}, \mathbf{A})$ for $\mathcal{X} \in\mathbb{R}^{N \times T}, \mathbf{A} \in \mathbb{R}^{N \times N}$, if $m$ new instances enter this set at some time t and the graph changes to $\mathcal{G}^{\prime} = (\mathcal{X}^{\prime}, \mathbf{A}^{\prime})$ for $\mathcal{X}^{\prime} \in\mathbb{R}^{(N + m) \times T}, \mathbf{A}^{\prime} \in \mathbb{R}^{(N+m) \times (N+m)}$. 

\begin{figure}[htbp]
\centerline{\includegraphics[width=0.49\textwidth]{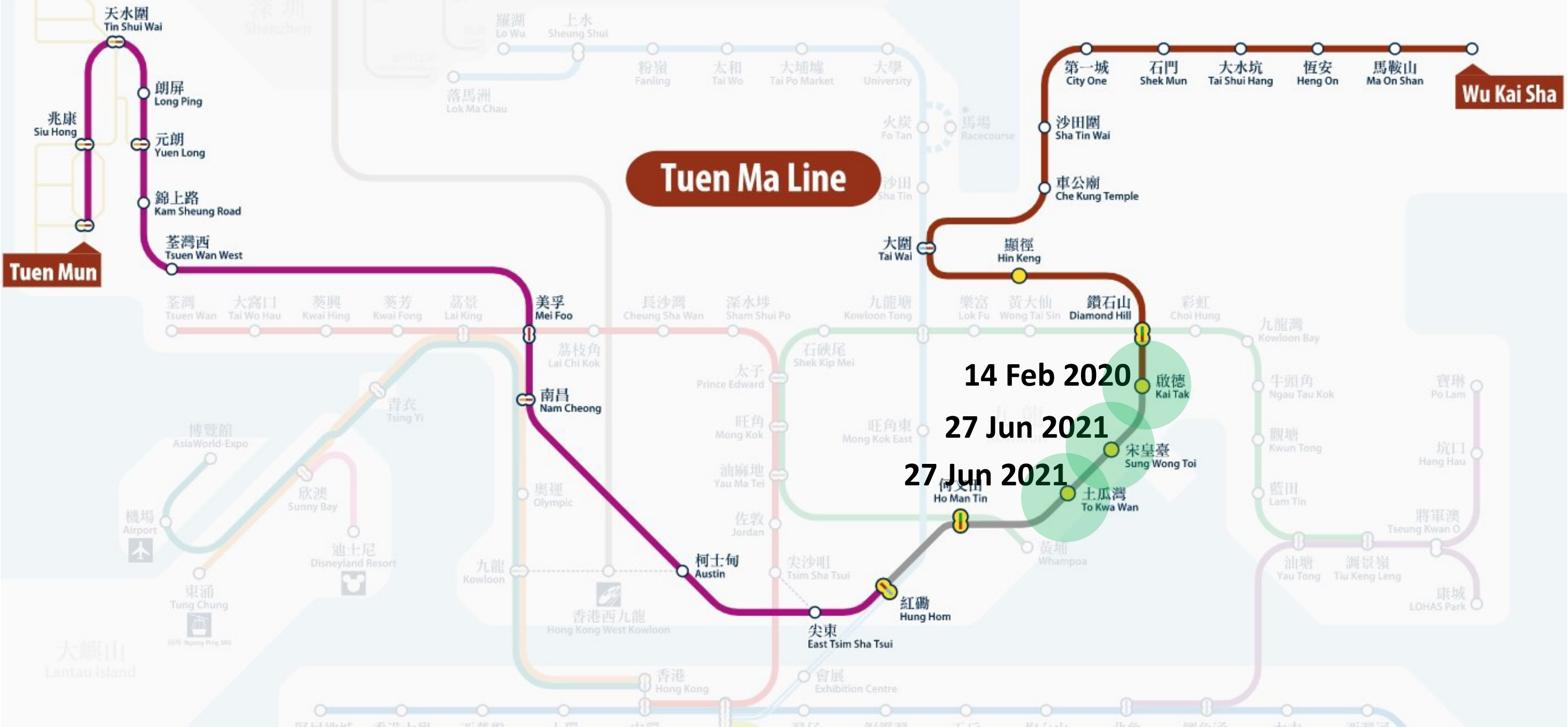}}
\caption{Three newly open stations (highlighted in green) have limited data due to cold-start. The three stations also connect two separate metro lines together. As a result, the graph structure also changed.}
\label{new}
\end{figure}

These newly joining $m$ instances have quite limited data available, which is known as the cold-start problem \cite{li2022profile}. The cold-start problem originated from the recommender system. In ITS,  the cold-start problem can also be quite common in reality. For example, as shown in Figure. \ref{new}, due to the construction of new metro stations or roads, there are new metro stations joining the metro network (on 14 Feb 2020, 27 Jun 2021, and 27 Jun 2021, respectively), and also simultaneously changing the network graph structure (for example, the new grey line connects the purple line and the brown line together). Since these new instances often have limited records, transferring knowledge from existing instances to these new instances is vital and challenging. In this task, we directly use the encoder and linear regressor learned on old data on new instances to forecast the future value, and we evaluate its performance on a real-world passenger inflow dataset, Hong Kong Metro, including data from 90 metro stations in the first three months of 2017 and 93 stations in June 2021. 3 new stations were opened between these years. The raw data is aggregated into 5 minutes, and we have 247 time steps in one day due to the operation time of the metro system.

\begin{figure*}[t]
\centerline{\includegraphics[width=0.99\textwidth]{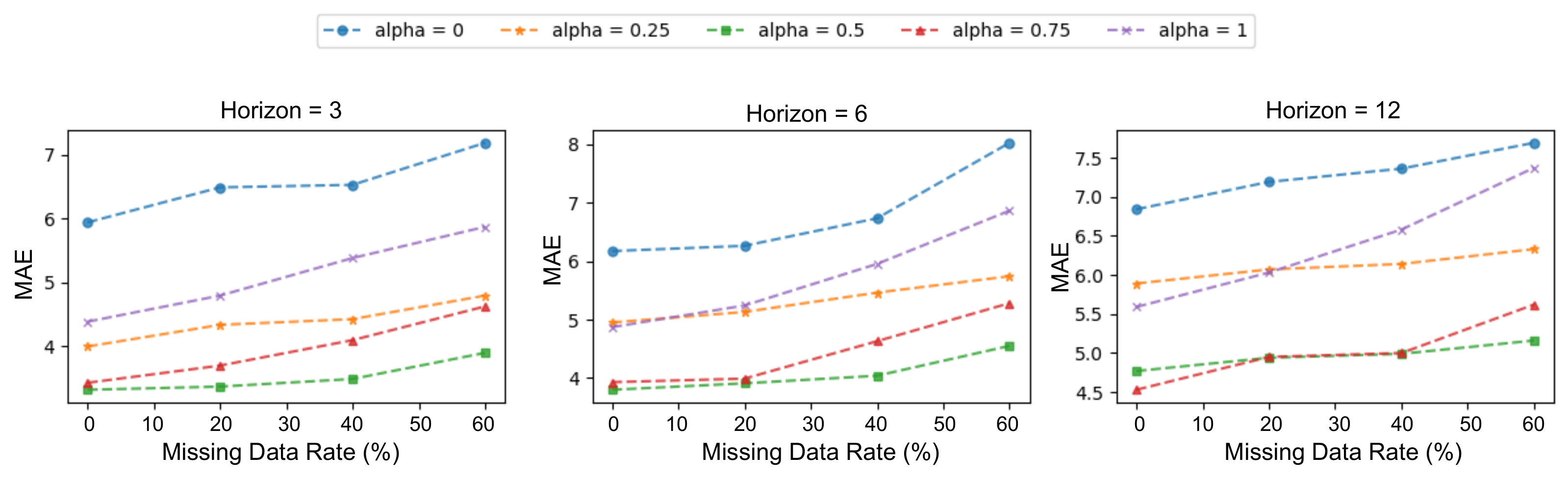}}
\caption{MAE of METR-LA dataset with respect to different missing data rate and $\alpha$}
\label{missing_data}
\end{figure*}

We learned the nonlinear embedding and linear regression model on data in 2017 and applied the encoders and regressor directly to the data in 2021 to forecast passenger inflow in the three new stations. Similarly, we apply RMSE, MAE, and MAPE to evaluate the forecasting performance and compare our method with the following baselines: 1) SVR, 2) FC-LSTM, and 3) TS2Vec. These baselines are trained on a single time series of data in 2017 to adapt to new stations.

\begin{table}[htbp]
	\centering
	\fontsize{7}{10}\selectfont    
	\caption{City Metro for new stations: MAE, MAPE and RMSE for proposed method and the baselines. The best results are bolded.}
	\begin{tabular}{ccccc}
		\toprule
		\toprule
		\multirow{2}{*}{Model}&
		\multicolumn{3}{c}{City Metro Dataset (15/30/60 min)} \cr
		\cmidrule(lr){2-4}
		& RMSE & MAE & MAPE(\%)  \cr
		\cmidrule(lr){1-4}

		SVR & 32.212/39.88/49.36 & 20.03/26.74/30.29 & 29.52/35.68/40.06 \cr
		FC-LSTM & 28.40/37.93/46.69 & 17.22/24.41/26.37 & 27.79/33.86/\textbf{39.61} \cr
		TS2Vec & 28.76/39.30/50.29 & 17.51/23.87/29.99 & 29.35/35.11/51.74 \cr
		Proposed & \textbf{24.02/31.15/46.33} & \textbf{14.55/17.85/25.12} & 
		\textbf{25.24/31.164}/46.21  \cr
		
		\bottomrule
		\bottomrule
	\end{tabular}\vspace{0cm}
	\label{new_station}
\end{table}

From Table \ref{new_station}, our method achieves the best forecasting performances in most of the cases. Especially for short-term forecasting, i.e., 15 min, our method reduces RMSE by 14$\%$, MAE by 19$\%$, and MAPE by 18$\%$ compared to the second-best model. Generally, STGNN models are SOTA in multivariate time series forecasting. However, they are not flexible to change of graph and may fail in many real-world cases. Note that for both downstream tasks, the representations only need to be learned once for different horizons with linear regression, which demonstrates the efficiency of our framework.

\section{Further Analysis}
\label{sec: analysis}
\subsection{Ablation Study}
To verify the effectiveness of the temporal and spatial components in our model, we compare the model performance with different values of the hyper-parameter $\alpha$ on the METR-LA dataset. We vary the value of $\alpha$ from 0 to 1 with the interval of 0.25, where $\alpha = 1$ means we only use historical information and $\alpha = 0$ we only consider spatial dependency. 

\begin{table}[htbp]
	\centering
	\fontsize{8}{11}\selectfont    
	\caption{Ablation Results on METR-LA with respect to diffenet value of $\alpha$. The best results are bolded.}
	\begin{tabular}{ccccc}
		\toprule
		\toprule
		\multirow{2}{*}{$\alpha$}&
		\multicolumn{3}{c}{METR-LA Dataset (15/30/60 min)} \cr
		\cmidrule(lr){2-4}
		& RMSE & MAE & MAPE(\%)  \cr
		\cmidrule(lr){1-4}
		
		0 & 8.98/10.32/11.98 & 5.94/6.17/6.84 & 10.10/10.35/14.76   \cr
		0.25 & 7.13/8.68/10.00 & 3.99/4.95/5.89 & 9.60/12.70/17.40 \cr
		0.5 & \textbf{6.63/7.51}/9.02 & \textbf{3.31/3.80}/4.77 & 
		\textbf{7.80/9.44}/12.23  \cr
		0.75 & 6.67/7.84/\textbf{8.73} & 3.42/3.93/\textbf{4.53} & 8.73/10.03/\textbf{12.00} \cr
	  1 & 8.12/9.23/10.30 & 4.38/4.87/5.59 & 
        9.89/10.77/14.80  \cr
		
		\bottomrule
		\bottomrule
	\end{tabular}\vspace{0cm}
	\label{ablation}
\end{table}

Table \ref{ablation} reveals that our model performs best for short-term forecasting when $\alpha$ = 0.5. When the temporal target contributes more than the spatial target, or when $\alpha$ = 0.75, the model performs better for long-term prediction. This is because, for relative long-term prediction, temporal information of individual instances containing historical patterns contributes more to our model. Furthermore, the performance of our approach clearly degrades for w/o temporal target or w/o spatial target.

\subsection{Robust to Missing Data}
Missing data is a common issue for time series collected from the real world. Our proposed model uses aggregated representations rather than original discrete data to train the forecast model, which demonstrates more stable performances on missing data. Intuitively, the network can infer the spatiotemporal representations with incomplete contexts due  to the design of timestamp masking. Assume that we have pre-processed data without missing values in the training stage, and test the model with missing data only in the forecasting stage. We take METR-LA dataset as an example to evaluate the forecasting performance with randomly missing data of three missing rates, 20$\%$, 40$\%$, and 60$\%$. Figure \ref{missing_data} shows that our method with both spatial and temporal targets is extremely robust under a missing rate of less than 60$\%$. Note that with smaller $\alpha$, e.g., $\alpha = 0.25$, the forecasting performance is more stable than that with larger $\alpha$, e.g., $\alpha = 0.75$. This indicates that the design of spatial targets could improve the robustness of our representation learning framework.

\section{Conclusion}
\label{sec: conclusion}
This paper proposes a time step-level self-supervised representation learning framework for correlated time series, which learns the embedding at specific time step $t$ for individual instances via using a masked view to predict the representation of the bootstrapped spatio-temporal targets. We evaluate our learning framework on two downstream tasks: CTS forecasting and cold-start transferring models to new instances with limited data. We show that our model has the best performance in most of the cases compared to baselines. Furthermore, we demonstrate the effectiveness of each proposed component in our model and its robustness to missing data. 

In the future, we will extend our framework to more datasets and downstream tasks, e.g., anomaly detection, to explore the effectiveness of our representation learning framework. 

\section*{ACKNOWLEDGMENT}

This paper was supported by grant FSUST20-FYTRI03B. 

\balance
\bibliographystyle{IEEEtran} 
\bibliography{main} 

\end{document}